\documentclass[letterpaper]{article} 
\usepackage{aaai24}  
\usepackage{times}  
\usepackage{helvet}  
\usepackage{courier}  
\usepackage[hyphens]{url}  
\usepackage{graphicx} 
\usepackage{natbib}  
\usepackage{caption} 
\usepackage{algorithm}

\usepackage{algorithmic}
\usepackage{amsmath}

\usepackage{newfloat}
\usepackage{listings}
\DeclareCaptionStyle{ruled}{labelfont=normalfont,labelsep=colon,strut=off} 
\floatstyle{ruled}
\newfloat{listing}{tb}{lst}{}
\floatname{listing}{Listing}

\title{\bf Reinforcement Learning from Statistical Feedback: the Journey from AB Testing to ANT Testing}
\author{Feiyang Han 
\textsuperscript{\rm 1,} 
\footnote{E-mail: 19110180030@fudan.edu.cn.}
\quad  Yimin Wei 
\textsuperscript{\rm 2,} 
\footnote{E-mail: ymwei@fudan.edu.cn and yimin.wei@gmail.com}
\quad Zhaofeng Liu 
\textsuperscript{\rm 3,}
\footnote{E-mail: zhaofengliu@tencent.com.}
\quad Yanxing Qi 
\textsuperscript{\rm 3,}
\footnote{E-mail: yanxingqi@tencent.com.}
}
\affiliations{
\textsuperscript{\rm 1} School of Mathematical Sciences, Fudan University, Shanghai, 200433, P. R.  China. This author is supported by the National Natural Science Foundation of China under grant 12271088. \\
\textsuperscript{\rm 2} Corresponding author. School of Mathematical Sciences and  Shanghai Key Laboratory of Contemporary Applied Mathematics, Fudan University, Shanghai, 200433, P. R. China. This author is supported by Innovation Program of Shanghai Municipal Education Commission and the National Natural Science Foundation of China under grant 12271088.\\
\textsuperscript{\rm 3}
Shenzhen Tencent computer System Co.Ltd., Shenzhen, Guangdong, 518054, P. R. China.
}

\usepackage{bibentry}

\begin{document}

\maketitle

\begin{abstract}
Reinforcement Learning from Human Feedback (RLHF) has played a crucial role in the success of large models such as ChatGPT. RLHF is a reinforcement learning framework which combines human feedback to improve learning effectiveness and performance. However, obtaining preferences feedback manually is quite expensive in commercial applications. Some statistical commercial indicators are usually more valuable and always ignored in RLHF. There exists a gap between commercial target and model training. In our research, we will attempt to fill this gap with statistical business feedback instead of human feedback, using AB testing which is a well-established statistical method. Reinforcement Learning from Statistical Feedback (RLSF) based on AB testing is proposed. Statistical inference methods are used to obtain preferences for training the reward network, which fine-tunes the pre-trained model in reinforcement learning framework, achieving greater business value. Furthermore, we extend AB testing with double selections at a single time-point to ANT testing with multiple selections at different feedback time points. Moreover, we design numerical experiences to validate the effectiveness of our algorithm framework.\\

\end{abstract}

\section{Introduction}
In recent years, the development of artificial intelligence technology has promoted the research and application of AI Generated Content (AIGC). From text generation to image generation, from music creation to video production, AIGC \cite{cao2023comprehensive} has covered numerous application scenarios. AI generated content can greatly improve production efficiency and reduce labor costs, while also maintaining quality and accuracy of content. With the emergence of ChatGPT and some image generation models, AIGC has entered people's lives and work with zero distance. Everyone can search for their own treasures in the ocean of AIGC. Meanwhile, it is undeniable that AIGC brings about revolutionary changes in business world, which generates plenty of new challenges and chances. In the training process of AIGC models, apart from the requirements of generating quality, logical correctness, and information completeness that meet human preferences, the differences in commercial value brought by different generation choices are also significant and cannot be ignored. The diverse data in commercial application scenarios not only pose significant challenges in data acquisition compared to human-labeled data but also hold greater commercial value. How to combine a unified business goal and AIGC model parameter learning will be an important research topic.

In addition to more advanced models and high-quality diverse training data, reinforcement Learning from Human Feedback (RLHF) is another important drivers for the success of AIGC models such as ChatGPT. In traditional reinforcement learning, the agent can only receive feedback through reward signals from the environment. However, in some cases, reward signals may be difficult to define or inaccurate. In such cases, human feedback can be used to supplement reward signals. The reinforcement learning framework combined with human expertise greatly expands the space for improvement of pre-trained models in the parameter fine-tuning stage. In reinforcement learning from human feedback, feedback data is usually in the form of scores or preferences given by humans, which might be very expensive to obtain. Meanwhile, in commercial applications, specific business indicators such as revenues and user retention rates are often more valuable, which are not token into consideration in RLHF. In this paper, we will focus on the gap between business objectives and neural network reinforcement learning.

A/B testing \cite{king2017designing, siroker2015b} is a method used to compare two different strategies, typically to determine which strategy is most effective for achieving a specific goal. In AB testing, the tested users are randomly divided into two groups: Group A and Group B \cite{gilotte2018offline}. The only difference between these two groups is the tested strategy, with other conditions fixed. Feedback results from both groups are then collected and compared to determine which strategy performs better statistically. It can be noticed that using AB testing to compare recommended business values can replace human preferences in RLHF. In this article, we use a reward network to predict such preferences, thus forming a reinforcement learning framework for business value feedback.

The main contributions of this research are summarized as follows,

\begin{itemize}
\item[(a)] Introducing Reinforcement Learning from Statistical Feedback (RLSF), an economical End-to-End framework with more business value considered, which utilizes the differences in commercial preferences generated by AB testing to fine-tune the pre-trained models through reinforcement learning.
\item[(b)] Promoting AB testing to AN testing, with a reward network to extend from binary choice preferences to n-choice preferences at one time point.
\item[(c)] Exploring ANT testing, which involves training different reward networks on multiple feedback cycles to fit different business indicators, with two reinforcement learning framework, gradual and one-time feedback.
\end{itemize}

This paper is organized as follows. In Section 2, some related researches are summarized and reviewed. Moreover, Reinforcement Learning from Business Feedback is introduced in Section 3 with statistical experiment details. Besides, in Section 4, we extend AB testing to AN testing and ANT testing, generating corresponding frameworks to fine-tune the pre-trained models. Numerical experiments are designed to illuminate the effectiveness of RLSF in Section 5. Finally, in Section 6, we summarize our theory and numerical results, obtaining the conclusions and future problems about our research.

\section{Related Works}

\subsection{Human Feedback Learning} 
Our study to obtain the business value instead of just improving generation quality is built on a lot of previous researches about Reinforcement Learning from Human Feedback \cite{christiano2017deep, ibarz2018reward, ouyang2022training}, such as human ratings or rankings. Moreover, reinforcement framework has been studied to fine-tune the pretrained NLP models to summarize text \cite{bohm2019better, gao2018april, stiennon2020learning, wu2021recursively, ziegler2019fine}. In addition, there are plenty of fields to train a reward model with human feedback and fine-tune with reinforcement framework \cite{biyik2022learning, madaan2022memory, sumers2021learning}. There are also lots of works to reduce the cost of tailored human feedback, such as \cite{liang2022reward}, which is also one of our motivations to take some statistical business preferences into consideration. For commercial Internet company, such as Google and Meta, there are a great deal of users and we do not need to label the preferences manually.

\subsection{Statistical Deep Learning} 
In 2017, there are researchers writing an important review about statistical, deep, and reinforcement learning \cite{sharma2017literature}, from which we have learned a lot. Besides, scholars compare different reinforcement learning algorithms \cite{colas2019hitchhiker}, especially introducing a hitchhiker's guide.
There are also plenty of researches, which focus on different fields about the combination of statistical inferences and deep learning \cite{bartlett2021deep, jin2021survey, shah2020feature, viharos2021reinforcement}. Our work also attempts to build the bridge between traditional statistical methods and reinforcement learning to achieve more commercial benefits.

\subsection{Deep Learning in Commercial Cases}
This part of deep learning is very attractive for researchers from not only company but also colleges. Taking an essential target, click-through rate (CTR), as an example, there are a great many researches to predict CTR in commercial cases \cite{wang2023cl4ctr, zhang2021deep}. Among them, deep learning approaches are utilized to obtain more information from users, such as deeper feature interaction \cite{guo2017deepfm} and attention block \cite{zhou2018deep}. Besides, other commercial indicators are also studied by deep learning method, such as Life Time Value \cite{chen2018customer} and retention index \cite{vrzal2021deeprei}. All these useful models and methods are also valuable in our RLSF framework, because all these strong well-trained reward networks could be utilized to predict these commercial indicators.

\section{RLSF Framework}

In order to incorporate business data indicators as feedback for reinforcement learning training, it is necessary to model the business data into feedback data that can be used for reinforcement learning. However, for individual users, binary indicators such as retention or continuous indicators such as in-App purchases are influenced by various objective or individual factors. Using them directly as feedback for reinforcement learning is not appropriate. Therefore, we utilize the results generated through statistical inference using AB testing with randomly sampled users as preference data that can provide business feedback.

\subsection{RLSF Framework with AB Testing}

Firstly, we need to choose the experimental objectives, such as retention rates of an application, pass rates of the next level in the game, or other business indicators, similar to traditional AB testing. Taking the example of an intelligent NPC in a text-based dialogue game scenario, we aim to facilitate user engagement and improve the retention rate in the game through interactions between users and NPCs. Additionally, we expect the NPC to serve as a guide for game level progression. Assume that the business indicators we are interested in for this experiment is $\eta$. 

Next, we will randomly select two groups of users with an equal number for AB testing. It is important to note that user sampling should be done in a way that makes the distribution of the two groups as similar as possible. For these two groups of users, we consider an agent interacting with an environment through a series of steps. We can model such a reinforcement learning scenario as follows: the agent takes an action $a$ based on the statement $s$ from the environment. For the two interaction flows that need to be tested, we can define them as $\mathcal F_1$ and $\mathcal F_2$, with their corresponding business indicators, $\eta_1$ and $\eta_2$. The hypothesis for this experiment is 
\begin{equation}
\begin{aligned}
H_0:\ \eta_1 = \eta_2,\ H_1:\ \eta_1\neq\eta_2,
\end{aligned}
\end{equation}
or 
\begin{equation}
\begin{aligned}
H_0:\ \eta_1 \leq \eta_2,\ H_1:\ \eta_1>\eta_2.
\end{aligned}
\end{equation}
 Of all the complete interaction flows, only one interaction produces a different sample, i.e.,
\begin{equation}
\begin{aligned}
\mathcal F_1=\left(\sigma_1,\cdots,\sigma_{n_0}^{(1)},\cdots,\sigma_{n}\right),\\ \ \mathcal F_2=\left(\sigma_1,\cdots,\sigma_{n_0}^{(2)},\cdots,\sigma_{n}\right),
\end{aligned}
\end{equation}
where 
$$
\begin{aligned}
\sigma_{n_0}^{(i)}=\left(\left(s_0^{(i)}, a_0^{(i)}\right), \ldots,\left(s_{k-1}^{(i)}, a_{k-1}^{(i)}\right)\right) \in\left(\mathcal S\times \mathcal A\right)^k,\\ i=1,2.
\end{aligned}
$$

Similarly, taking the example of an intelligent NPC in a text-based dialogue game scenario, the dialogue flow between the two experimental groups differs only in one round of dialogue. In the different round, each step's state consists of existing text tokens, while the corresponding set of potential actions comprises the entire word corpus. The next state is generated by sampling from the corpus based on the probabilities provided by the pre-trained model.

In this way, hypothesis testing can be conducted on the business indicators of the two user sample groups, considering the interaction flow that occurs only once. If the confidence level is below $\alpha$ and the statistical power is below $1-\beta$, the sample group is discarded. Otherwise, the relative commercial value ranking between the two sample groups is determined based on their relative sizes of $\sigma_1$ and $\sigma_2$.

\textbf{Definition 1.}
We could define that the preferences $\succ$ in $\left(\mathcal{S} \times \mathcal{A} \right)^k$ if
$$
\begin{aligned}
\sigma_1=\left(\left(s_0^1, a_0^1\right), \ldots,\left(s_{k-1}^1, a_{k-1}^1\right)\right)\succ\\ \left(\left(s_0^2, a_0^2\right),\ldots,\left(s_{k-1}^2, a_{k-1}^2\right)\right)=\sigma_2
\end{aligned}
$$
whenever the business feedback $\eta_1$ and $\eta_2$ satisfies that $\eta_1>\eta_2$.

Next, we need to provide such business value preferences as feedback to the pre-trained model. Similar to human feedback reinforcement learning, we use neural networks to learn these preferences. Specifically, we train a suitable reward network $r$ which satisfies that
$$
\begin{aligned}
r\left(s_0^1, a_0^1\right)+\cdots+r\left(s_{k-1}^1, a_{k-1}^1\right)\\
>r\left(s_0^2, a_0^2\right)+\cdots+r\left(s_{k-1}^2, a_{k-1}^2\right)
\end{aligned}
$$
whenever $\sigma_1\succ\sigma_2$. Furthermore, the reward network is trained to minimize the cross-entropy loss between the predicted preference distribution and the actual statistical business preference distribution, i.e.,
\begin{equation}
\begin{gathered}
\begin{aligned}
\operatorname{loss}(\hat{r})=-\sum_{\left(\sigma_1, \sigma_2, \delta\right) \in \mathcal{D}} \kappa(\sigma_1, \sigma_2)\log \hat{P}\left[\sigma_1 \succ \sigma_2\right]\\ +\kappa(\sigma_1, \sigma_2) \log \hat{P}\left[\sigma_2 \succ \sigma_1\right]
\end{aligned}
\end{gathered}
\end{equation}
where 
$$
\hat{P}\left[\sigma_1 \succ \sigma_2\right]=\frac{\exp \sum \hat{r}\left(s_t^1, a_t^1\right)}{\exp \sum \hat{r}\left(s_t^1, a_t^1\right)+\exp \sum \hat{r}\left(s_t^2, a_t^2\right)} 
$$
is the predicted preference distribution and $\delta$ is the actual statistical business preference distribution from AB testing, i.e.,
\begin{equation}
\kappa(\sigma_1,\sigma_2)=
\begin{cases}
\delta^{(1)},\ \ if \ \ \sigma_1 > \sigma_2,\\
\delta^{(2)},\ \ if \ \ \sigma_1 < \sigma_2,\\
\left[\delta^{(1)}+\delta^{(2)}\right]/2,\ \ if \ \ \sigma_1 = \sigma_2.
\end{cases}
\end{equation}
Here $\delta^{(x_0)}$ is the one point distribution at $x_0$, which means that $\delta^{(x_0)}(x)=1$ if and only if $x = x_0$.

\subsection{Some Details in AB testing}

In AB testing, we first need to specify the significance level $\alpha$, the type II error $\beta$, and the sample size $M$. Besides, we need to select $M$ samples without replacement as unbiasedly as possible to form two sample groups to test the preference differences between two choices generated by our pretrained model. 

\begin{figure*}[h]
\centering
\includegraphics[width=0.9\textwidth]{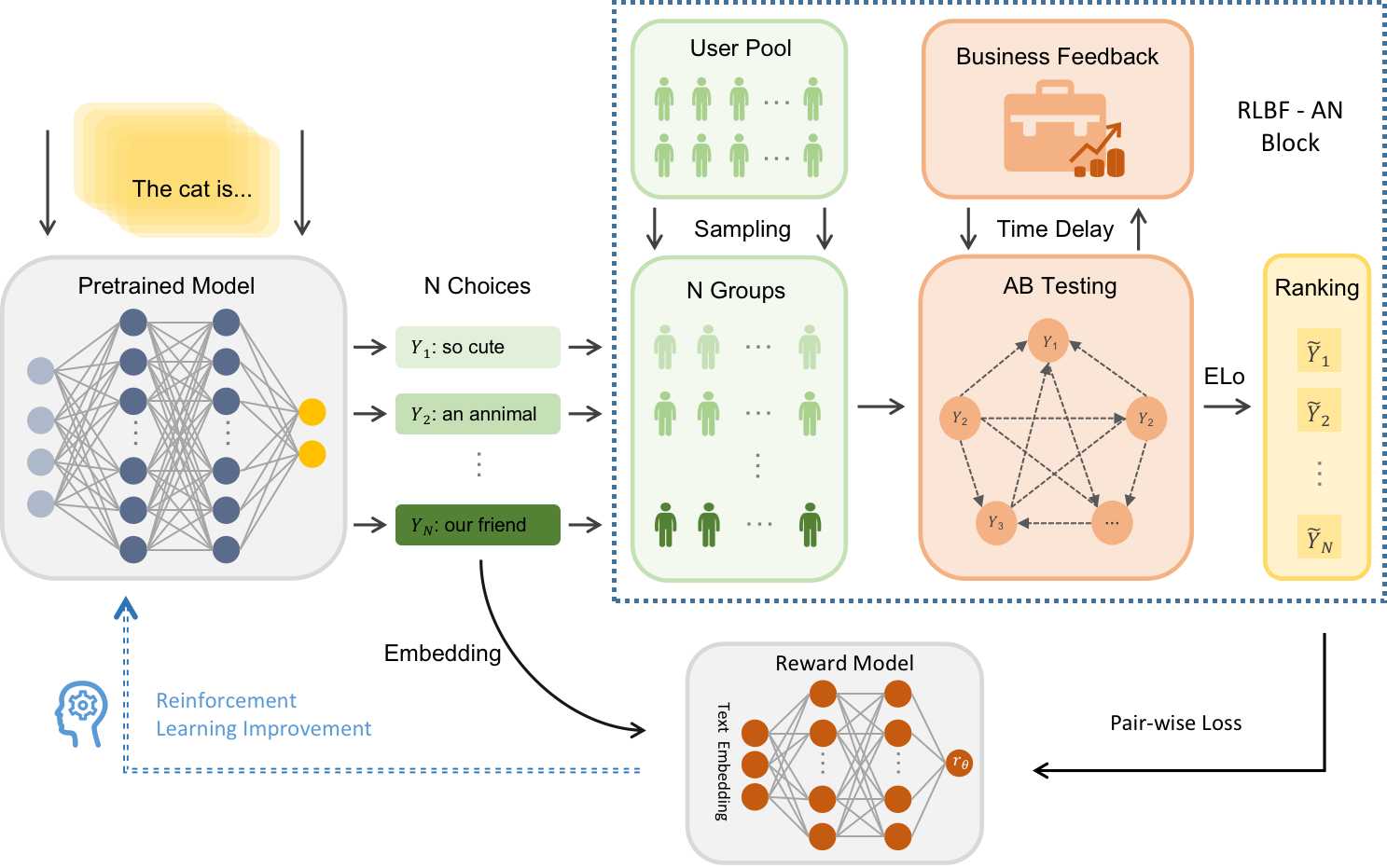} 
\caption{Reinforcement Learning Framework from Statistical Feedback based on AN Testing. For taking AB testing among N choices generated by the pretrained model, the users sampled from user pool should be divided into N groups. After AB testing between any two user groups, we can obtain $C_N^2$ statistical business preference feedback, which could take the choices in order with ELo Algorithm \cite{zanco2022comprehensive}. Furthermore, the rank trains the reward model, minimizing the pair-wise loss. Finally, the reward model can fine tune the pretrained model, where there are plenty of reinforcement algorithms to be utilized.}
\label{fig1}
\end{figure*}

Then, we need to determine whether there is a significant difference in the business indicators between the two experimental groups. Without loss of generality, we assume $\eta_1 > \eta_2$. At this time, the null hypothesis and alternative hypothesis are

\begin{equation}
\label{H_1}
H_0:\ \eta_1 = \eta_2,\ H_1:\ \eta_1\neq\eta_2.
\end{equation}

There may be some invalid points in the samples and we should remove these invalid points. Suppose that the rest sample sizes of two groups are $M_1$ and $M_2$, which satisfies $M_2=k\cdot M_1$. The t-statistic can be expressed as:
\begin{equation}
\label{t-value}
t=\frac{\bar X_1 - \bar X_2}{S_p\sqrt{\frac{1}{M_1} +  \frac{1}{M_1}}},
\end{equation}
where
\begin{equation}
 S_p^2=\frac{(M_1-1)S_1^2+(M_2-1)S_1^2}{M_1+M_2-2}.
\end{equation}
If $t<t_{1-\alpha/2}$, we accept the null hypothesis. Otherwise, we reject the null hypothesis and choose the alternative hypothesis.

Furthermore, we need to test whether there is a significant difference between these two samples. The target gap $\delta$ to be tested should satisfy that $\delta=\max\{\delta_0, |\eta_1-\eta_2|\}$, where $\eta_i$ is the indicator from sample $i$ and $\delta_0$ is the minimum target gap given at first. Here we rename $\eta_i$ as $\mu_i$ for convenience. At this step, the null hypothesis and alternative hypothesis are

\begin{equation}
\label{H_2}
H_0:\ \mu_1 - \mu_2=\delta,\ H_1:\ \mu_1 - \mu_2\neq \delta.
\end{equation}

According to traditional statistical knowledge, the minimum sample size for testing mean-related indicators, such as average playback duration, average payment per user, etc., can be calculated from
\begin{equation}
\label{N_1}
n_1=\frac{(\sigma_1^2+\sigma_2^2/k)(z_{1-\alpha/2}+z_{1-\beta})^2}{\delta^2}
\end{equation}
and $n_2=k\cdot n_1,$ where $n_i$ is the minimum size, $\mu_i$ is the mean of target business value and $\sigma_i$ is the standard deviation of target business value of $i$-th sample group, for $i=1,2$, respectively. Moreover, $z_{1-\alpha/2}$ and $z_{1-\beta}$ could be found from the normal distribution table. 

However, when testing ratio-related indicators, such as click-through rate, 7-day retention rate, etc., the minimum sample size is
\begin{equation}
\label{N_2}
n_1=\frac{\left[\sqrt{\bar{p} \bar{q}\left(1+\frac{1}{k}\right)} z_{1-\alpha / 2}+\sqrt{p_1 q_1+\frac{p_2 q_2}{k}} z_{1-\beta}\right]^2}{\delta^2} 
\end{equation}
and $n_2=k\cdot n_1,$ where $p_i$ is the target business value of $i$-th sample group, for $i=1,2$, respectively, and
\begin{equation}
\begin{aligned}
&q_1=1-p_1,\ q_2=1-p_2, \ \ \\
&\bar p=\frac{p_1+kp_2}{1+k},\ \ \bar q= 1-\bar p.
\end{aligned}
\end{equation}
Similarly, $\eta_i$ is renamed as $p_i$ in this case.

If the sample sizes of the two groups, $M_1$ and $M_2$, respectively, satisfy that  
\begin{equation}
n_1\leq M_1,\ \ n_2\leq M_2,
\end{equation}
we will accept the null hypothesis that there is significant difference in the business indicators between the samples. If $\delta > 0$, we can gain the result that the first choice is better, while the second choice is determined to be better, if $\delta < 0$. Otherwise, there are two choices. The first choice is that we reject the null hypothesis and conclude that there is no significant difference in the business indicators between the samples. However, the other choice is to sample more users from user pool to build two bigger test groups and take the above AB testing again, where the minimum sizes $n_1$ and $n_2$ could be referred. The preference from the set with 4 elements, $\{$ $\eta_1 > \eta_2$, $\eta_1 < \eta_2$, $\eta_1 = \eta_2$, no result$\}$, can be utilized to train a reward model, with no need to cost a lot for labeling preference data manually like original RLHF.

The algorithm framework is summarized in the next page.

\begin{algorithm}
\caption{RLSF with AB Testing}
\begin{algorithmic}
\REQUIRE Target indicator, $\eta$; Target indicator gap, $\delta_0$; Target sample size, $M$; two Choices, $\sigma_1$ and $\sigma_2$; the significance level, $\alpha$; the type II error, $\beta$.
\ENSURE the Preference between two choices.
\STATE\textbf{Step 1. } Sample two user groups with $M$ members from user pool and offer different flows to these groups. Remove invalid points and get the sample sizes, $M_1$ and $M_2$.
\STATE\textbf{Step 2. } Take T-test on the hypothesis (\ref{H_1}). Compute the t-value from (\ref{t-value}) to determine whether there exists a significant difference between two groups.

\IF {$\eta_1=\eta_2$}
    \RETURN $\sigma_1=\sigma_2$
\ELSE
\STATE Test the hypothesis (\ref{H_2}). Compute the minimum sizes of samples, $n_1$ and $n_2$, from (\ref{N_1}) or (\ref{N_2}).
    \IF{$n_1\leq M_1 $ and $ n_2\leq M_2$}
        \IF{$\delta >0$}
            \RETURN $\sigma_1\succ\sigma_2$
        \ELSE
            \RETURN $\sigma_2\succ\sigma_1$
        \ENDIF
    \ELSE
        \RETURN no result \\
        \textbf{Or} sample more users and back to \textbf{Step 1. } (\textbf{$\star$})
    \ENDIF
\ENDIF
\end{algorithmic}
\end{algorithm}

\section{Extensions from AB Testing to ANT Testing}
In this section, we will generate a framework to solve the problem about how to handle more than two choices in one reinforcement feedback step. Moreover, the cases with two or more steps are also considered.

\subsection{RLBF Framework with AN Testing}

In the last section, we derive the RLBF (Reinforcement Learning framework from Business Feedback) based on AB testing. Furthermore, if there are more than two choices for a fixed position interaction within the same interaction flow, we need to construct a more generalized AN testing framework. For example, in the context of text-based dialogue, it can be used to determine the relative commercial value of potential answers to a specific question. Similarly, in the field of image generation, such as AI-generated advertising images, the relative click-through rates of various choices in a recommended and ranked stream of materials can be evaluated. We refer to this training process as Reinforcement Learning from Business Feedback based on AN Testing, emphasizing the influences of N choices instead of just the impact of two choices in AB testing on commercial indicators.

\begin{figure*}[h]
\centering
\includegraphics[width=0.9\textwidth]{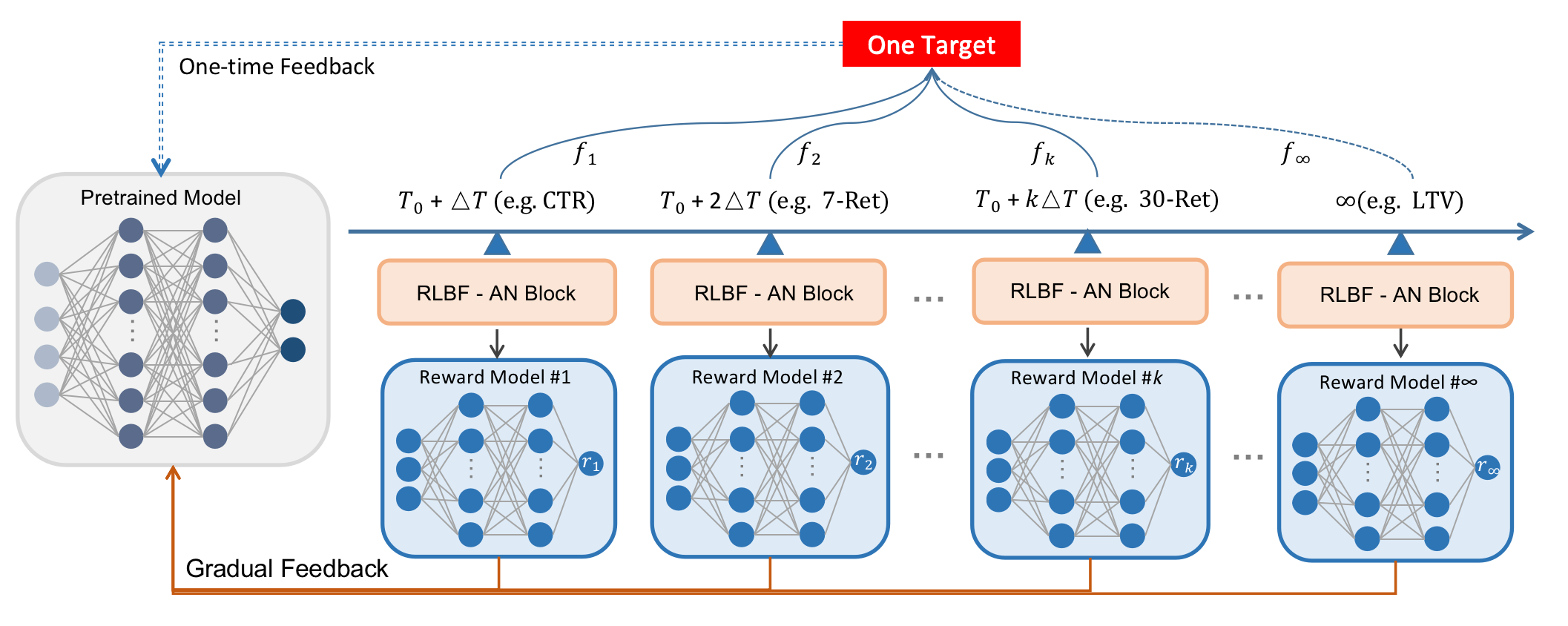} 
\caption{Reinforcement Learning Framework from Statistical Feedback based on ANT Testing. CTR means the Click Through Rate, 7(30)-Ret means the 7(30) day Retention and LTV means the Life Time Value. Some immediate indicators like CTR could be gained soon and plenty of indicators, for example 7(30)-Ret, are all time-delaying, while some indicators such as LTV can not be computed and should be predicted. In this time series, we could take the reinforcement feedbacks one by one, naming as the gradual feedback. Besides, we can build a functional to consider all indicators together, not only obtatined but also predicted, into one target to fine tune, naming as one-time feedback.}
\label{fig2}
\end{figure*}

Taking text generation as an example, it can be modeled as a sequential decision-making problem in the token space. Firstly, the context is treated as the state in reinforcement learning, and the action is selecting the next token after the previous token. In this case, there are N choices available. To meet the requirements of AN testing, N groups of user pools need to be selected, and AB testing is conducted between each pair. Based on this AB testing, using the methods discussed in the previous chapter, we can obtain $C_N^2$ statistical business feedback preferences.  The figure below illustrates the RLBF framework based on AN testing in the field of text generation. The direction of the arrows between the nodes in AB testing block from the diagram represents the direction of the preference order, while the absence of arrows indicates that there is no significant difference in the statistical business feedback indicators between the two choices. Moreover, we can utilize the Elo algorithm to establish a complete ranking among different choices within the same prompt. In this way, we can use a pair-wise loss function to fit such ranking results. As a result, we obtain scores for responses corresponding to the same prompt, which will be used for reinforcement learning of the pretrained model.

From the above analysis, it should be noticed that AB testing is a special case of AN testing, which allows for more flexibility in handling multi-choice problems and capturing differences between different choices. Figure (\ref{fig1}) presents the whole framework of Reinforcement Learning from Statistical Feedback based on AN testing.

\subsection{RLBF Framework with ANT Testing}
This section discusses the RLBF framework based on ANT testing with multiple feedback cycles and multiple choices, which could be referred in Figure \ref{fig2}. Commercial data feedback cycles naturally have asynchronous characteristics. For example, important commercial data includes conversation completion rate obtained immediately, next-day retention feedback delayed by one day, monthly active users delayed by one month, and lifetime value (LTV) feedback similar to an infinite horizon. We need to train reward networks separately for each of these feedback cycles.

Regarding the fusion of multiple rewards, we design two reinforcement learning modes, incremental feedback and one-time feedback. The incremental feedback mode means updating the reinforcement learning after each commercial data feedback. This mode has the advantage of simple operation and no additional computational overhead. However, the disadvantage is also relatively obvious, as the segmented reward function makes it difficult for the pre-trained model to achieve the optimal expected result.

The one-time feedback mode requires the fusion of multiple reward functions, using a functional or some neural networks.
\begin{equation}
r_{target}=\mathcal F\left(\sum_{k=1}^\infty w_k f_k(r_k)\right)
\end{equation}
 In addition, unreceived feedback (such as LTV) needs to be predicted. This mode enables flexible selection and creation of more valued commercial indicators by adjusting the weight of multi-objective fusion. For example, if the weight of a commercial indicator is set to approach infinity, it is consistent with the commonly used ``North Star Indicator" in traditional business analysis. However, the disadvantage of one-time feedback is also prominent, as it requires higher expert prior knowledge, and unreceived feedbacks need to be predicted, which brings significant computational overhead and errors.

\section{Numerical Experiements}
We test our RLSF framework on the fine-tuning process of pre-trained comment generation network based on Transformer structure. Our purpose is to attract users to click on the detailed page of the products through better comments. In this case, labeling manually by virtual users is meaningless and might be harmful to ignore the true preferences. However, statistical results can teach the pre-trained network better with nearly no cost, where PPO algorithm \cite{engstrom2020implementation, schulman2017proximal} is adopted. We use the emotional rating network to present a score for the comments generated by pre-trained model. From the following figure, the pre-trained model can generate the comments with more stable and better quality.

\begin{figure}[h]
\centering
\includegraphics[width=0.9\columnwidth]{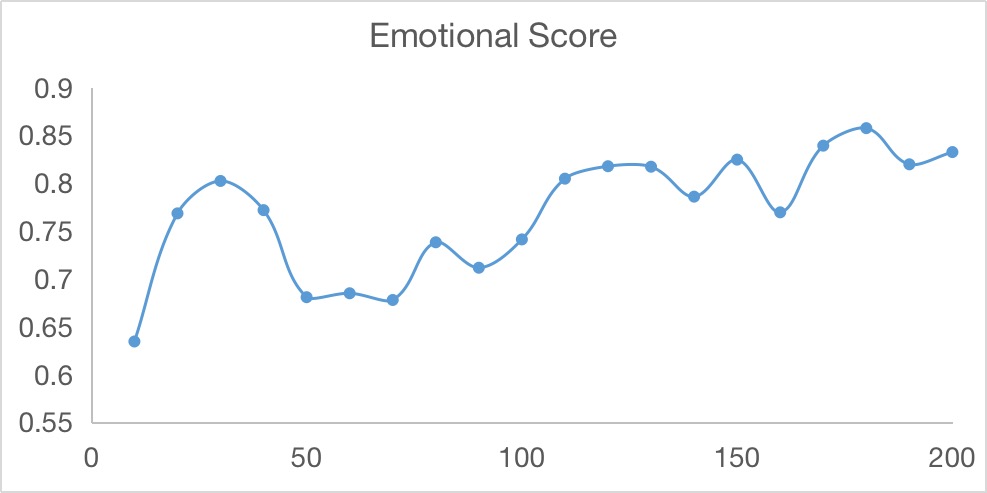} 
\caption{Emotional scores during the Training process, where the horizontal axis is the epochs. The emotional scores become more and more stable during the fine-tuning process.}
\label{fig1}
\end{figure}

\begin{figure}[h]
\centering
\includegraphics[width=0.9\columnwidth]{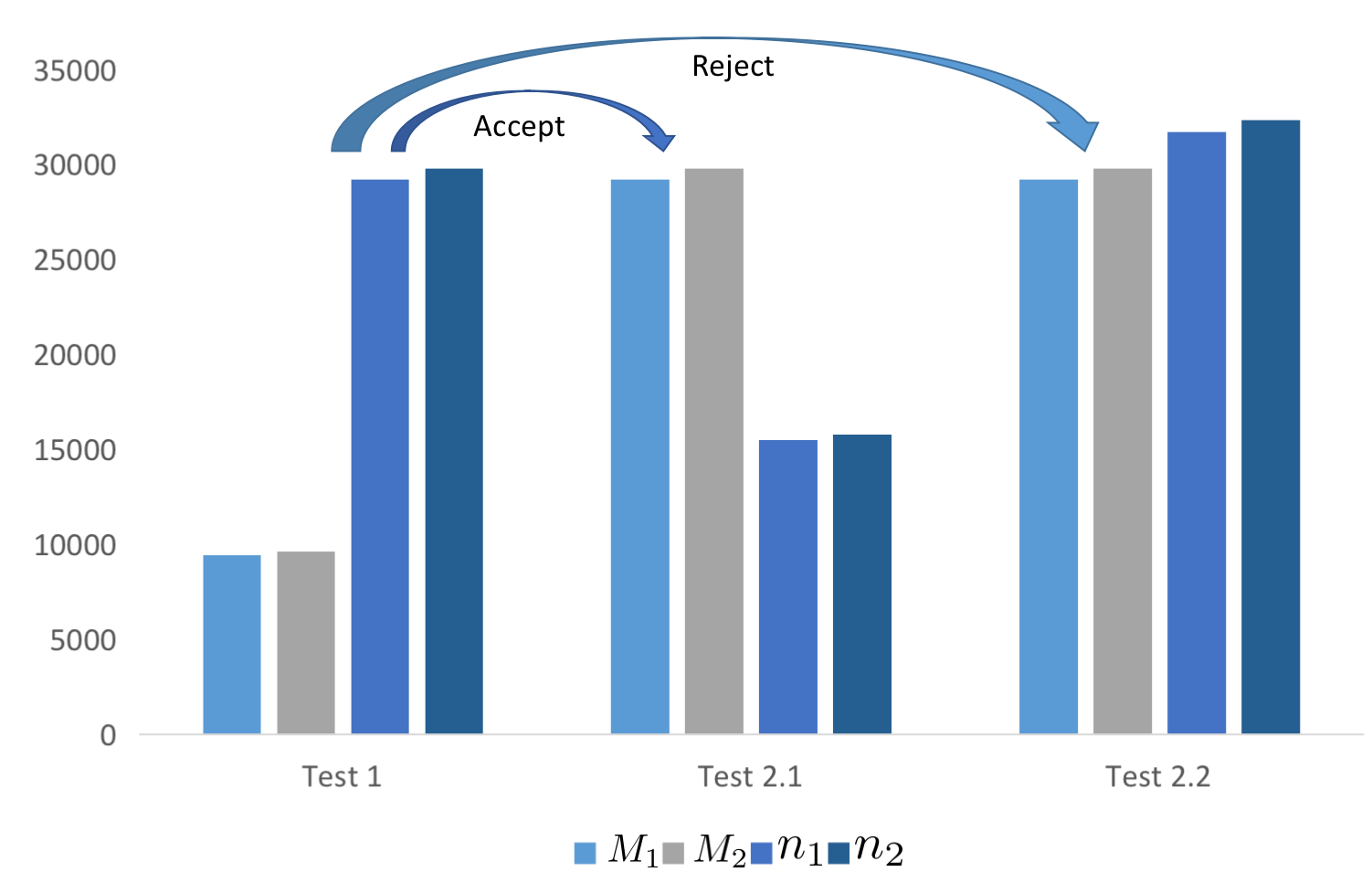} 
\caption{After Test 1, $n_i\leq M_i,i=1,2$, are not satisfied. After resampling, we retest them again. We accept the zero hypothesis if $n_i\leq M_i,i=1,2$ like Test 2.1, while the zero hypothesis should be rejected in Test 2.2.}
\label{fig1}
\end{figure}

In the above experiment, we set the minimum target increase as $\delta_0=1\%$, the confidence level as $\alpha=0.05$ and the statistical power as $\beta=0.2$. RLSF framework indeed improve the Click-Through Rate than the initial pre-trained model, which only considers the accuracy of generated texts but no commercial indicators. 

Besides, we test the minimum sample sizes and the resample framework in AB testing, which could be found at ($\star$) in Algorithm 1.

\section{Summaries}
In this paper, we propose a Reinforcement Learning from Statistical Feedback (RLSF) based on hypothesis testing, which reduces the reliance on expensive manually labeled preference data, while also taking the statistical business preferences into consideration to obtain more commercial value. In addition to AB testing, we also introduce AN testing, combined with ELo sorting algorithm to solve the problems with more than two choices at one time. Moreover, ANT testing is  considered to handle time series feedback, including not only time delaying but some predicted feedback. Finally, we design an experiment to illuminate the effectiveness of RLSF framework.

Looking to the future, we believe that this framework can be used not only in the field of text generation, but also in the field of image or audio generation, which might bring greater academic and practical value and attract more scholars to research on this topic.

\bibliography{aaai24}

\end{document}